\documentclass[letterpaper, 10 pt, journal, twoside]{IEEEtran}
\usepackage{graphicx}
\usepackage{bm}
\usepackage{amsmath}
\usepackage{dsfont}
\usepackage{mathtools}
\usepackage{amsfonts}
\usepackage{algorithm}
\usepackage{algpseudocode}	
\usepackage{url}
\usepackage{booktabs} 
\usepackage{hyperref}
\usepackage{color}  
\usepackage{multirow}
\usepackage{array}
\usepackage{epstopdf}
\renewcommand{\thefigure}{\arabic{figure}}
\title{A Visual Reinforcement Learning-Based Separate Primitive Policy for Peg-in-Hole Tasks}

\author{Zichun Xu, Zhaomin Wang, Yuntao Li, Lei Zhuang,~\IEEEmembership{Graduate Student Member,~IEEE}, Zhiyuan Zhao, \\
Guocai Yang, and Jingdong Zhao,~\IEEEmembership{Senior Member, IEEE}
\thanks{Received 26 August 2025; accepted 2 January 2026. Date of publication 19 January 2026; date of current version 16 February 2026. This article was recommended for publication by Associate Editor A. Wagenmaker and Editor A. Faust upon evaluation of the reviewers’ comments. This work was supported by the National Natural Science Foundation of China under Grant T2388101, Grant92148203,and Grant62403171. (\emph{Corresponding author: Jingdong Zhao})}
\thanks{Zichun Xu, Lei Zhuang, Guocai Yang, and Jingdong Zhao are with the State Key Laboratory of Robotics and Systems, Harbin Institute of Technology, Harbin 150001, China. (e-mail: zhaojingdong@hit.edu.cn)}
\thanks{Zhaomin Wang is with the Ubtech Robotics, Shenzhen 518000, China.}
\thanks{Yuntao Li is with the Meituan Academy of Robotics, Shenzhen 518000, China.}
\thanks{Zhiyuan Zhao is with the School of Mechanical Engineering, Shandong University, Jinan 250061, China.} 
\thanks{This article has supplementary downloadable material available at https://doi.org/10.1109/LRA.2026.3655305, provided by the authors.}
\thanks{Digital Object Identiﬁer 10.1109/LRA.2026.3655305}
}

\begin{document}

\markboth{IEEE Robotics and Automation Letters. VOL.11, NO 3, March, 2026}%
{XU \MakeLowercase{\textit{et al.}}: VISUAL REINFORCEMENT LEARNING-BASED SEPARATE PRIMITIVE POLICY FOR PEG-IN-HOLE TASKS}

\maketitle

\begin{abstract}

For peg-in-hole tasks, humans rely on binocular visual perception to locate the peg above the hole surface and then proceed with insertion. This paper draws insights from this behavior to enable agents to learn efficient assembly strategies through visual reinforcement learning. Hence, we propose a \textbf{S}eparate \textbf{P}rimitive \textbf{P}olicy (S2P) to learn how to derive location and insertion actions simultaneously. S2P is compatible with model-free reinforcement learning algorithms. Ten insertion tasks featuring different polygons are developed as benchmarks for evaluations. Simulation experiments show that S2P can boost the sample efficiency and success rate even with force constraints. Real-world experiments are also performed to verify the feasibility of S2P. Ablations are finally given to discuss the generalizability of S2P and some factors that affect its performance.

\end{abstract}


\section{INTRODUCTION}
\IEEEPARstart{P}{eg-in-hole} is a popular but challenging assembly scenario, where slight misalignment can lead to failure. Narrow clearances and frequent contacts make traditional motion planning methods inefficient and increase unnecessary manual calibration costs. With its learning and robust adaptation capabilities, reinforcement learning (RL) can continuously refine the output action and adapt to changing conditions through trial and error. The usual modalities, such as vision \cite{yuan2022pretrained}, force/torque \cite{lee2024polyfit}, and tactile \cite{sferrazza2024power}, can be provided as observations to the agent to infer appropriate actions. Considering the hardware conditions, sensor noise, and the inability to obtain the pose information of holes in real scenarios, visual modality can be accessed at a lower cost and is a reasonable choice. Thus, visual RL has been extensively explored in recent years, primarily involving generalization and sample efficiency \cite{hansen2022pretraining, yuan2024learning}.

However, since the RL process relies heavily on extensive transitions, it raises higher demand on extracting valuable information from high-dimensional observations with fewer computational resources. Data augmentation \cite{hansen2021stabilizing} can effectively improve the sample efficiency of model-free RL algorithms. Moreover, policy design can be leveraged to further accelerate the insertion progress. Previous studies \cite{nasiriany2022augmenting, vuong2021learning} trained hierarchical RL policies to explore various combinations of different primitives under the assumption of known state information, which is idealistic and disregards the challenges of extracting valuable information with noisy observations. Analyzing the human insertion behavior, the entire sequence can be roughly interpreted as the mix of location and insertion. When insertion fails, subsequent adjustments are performed to relocate and reinsert, which will be repeated until success. Hence, excessive definition of primitives is inefficient for accomplishing peg-in-hole tasks. In addition, training with visual modality is inherently high-cost compared to state information, which is exacerbated when additional resources are required to train the high-level policy for hierarchical RL.

Inspired by the above, this paper presents a \underline{\textbf{S}}eparate \underline{\textbf{P}}rimitive \underline{\textbf{P}}olicy (S2P) to accelerate the visual RL progress. Two separate policies are trained simultaneously to derive location and insertion actions, respectively, which are executed sequentially by the agent. Ensemble Q-functions \cite{chen2021randomized} are utilized to estimate Q-values for separate actions, and a higher update-to-data (UTD) \cite{ma2024revisiting} ratio is set to further enhance the sample efficiency. The proposed framework is adapted to model-free visual RL algorithms. Many insertion tasks, including different shapes of pegs and holes, are designed to verify the effectiveness of this strategy. Simulation results demonstrate that S2P can accelerate the insertion progress with higher success rates. The feasibility of S2P is also validated with the real-world experiments. The main contributions of this paper are as follows:

\begin{enumerate}
	\item We propose a separate primitive policy based on the model-free visual reinforcement learning framework for peg-in-hole tasks.
	\item The proposed method can effectively accelerate the sample efficiency and success rate even with strict force constraints.
	\item A suite of peg-in-hole tasks integrated in simulation and the real-world experiment validate the effectiveness of our method.
\end{enumerate}

\section{RELATED WORK}
\textbf{Reinforcement Learning for Peg-in-Hole.} RL can automate the repetitive insertion process and realize compliance through impedance control \cite{kozlovsky2022reinforcement}. The target objects span from common primitives to industrial connectors \cite{kimble2020benchmarking}. In recent years, different modalities have been fused to extract more features for complex insertion tasks with tight clearance \cite{chen2024multimodality, jin2024visualforcetactile}. Tactile \cite{sferrazza2024power, dong2021tactilerl} and force/torque \cite{luo2019reinforcement} can sense more contact details and allow for higher pose uncertainty than that offered by vision. However, training and tuning in the real platform is impractical, and thus verification in simulation plays an important role. For the simulation results to be instructive, the discrepancies in the presentation of modality data and physical properties between simulation and reality must be considered. Additional methods are required to align the native simulator signals with real sensor performance, as the inherent noise causes many transition states to be missed during training \cite{akinola2025tacsl}. In contrast, vision and robot proprioception are low-cost and easy to access \cite{yasutomi2023visual}. Visual RL has been proven to be effective to accomplish insertion tasks \cite{shi2021proactive, tang2024automate}. Even from the multimodal perspective, visual is an essential component, which is our intention to accelerate the visual RL process.

\textbf{Visual Reinforcement Learning.} Visual RL has shown great promise in dealing with complex manipulation tasks \cite{ze2023hindex}. Due to the complex dynamics modeling of model-based algorithms, the majority of visual RL research has been conducted upon model-free algorithms. Improving sample efficiency and generalizability is fundamental in the visual RL domain \cite{bertoin2022look}, and data augmentation serves as a pivot to bridge these two areas. Image-level data augmentation intuitively improves the data efficiency by expanding the diversity of observation samples \cite{laskin2020reinforcement}. The benefits of data augmentation can be further exploited by underlying algorithms, like DrQ \cite{yarats2021image} and its successor DrQ-v2 \cite{yarats2022mastering}, to enhance sample efficiency. Some additional recipes are employed to reduce the overestimation of Q-value and stabilize the training process \cite{hansen2021stabilizing, almuzairee2024recipe}. In addition, a visual encoder pre-trained on out-of-domain datasets \cite{yuan2022pretrained, nair2022rm} and well-designed contrastive learning schemes \cite{yuan2024learning, zheng2023taco} can achieve competitive generalization performance. These impressive results shed light on visual RL, but there are still challenges for making decisions in complex scenarios. Unlike the above studies, we focus on the insertion strategy design to accelerate the sample efficiency.

\textbf{Motion primitives.} Motion primitives are basic reusable units and exhibit distinct behaviors from each other. Reasonable combinations of motion primitives can accomplish complex tasks. The implementation of motion primitives in the RL field is mainly through hierarchical policy \cite{nasiriany2022augmenting} and parameterized action spaces \cite{zhang2022learning}. For hierarchical RL, a high-level policy acts as a commander to learn how to rationally invoke different primitives generated by low-level policies. The parameterized action space can be discrete labels representing different primitives and even include additional primitive parameters. Motion primitives enhance the interpretability of behaviors, lowering the exploration complexity for the agent and leading to fast convergence. However, numerous predefined primitives might be required to meet the task requirements, and prior knowledge is indispensable \cite{vuong2021learning}. Furthermore, some previous works are conducted under restricted assumptions, like known pose information of objects \cite{marzari2021towards, lee2022hierarchical}. The environment state can only be gathered by external sensors. Hence, visual information is provided to the policy to infer different primitives in our work.

\section{PRELIMINARIES}
Visual RL can be formulated as an infinite-horizon Partially Observable Markov Decision Process (POMDP) \cite{bellman1957markovian} with the tuple $\left\langle \mathcal{O}, \mathcal{A}, \mathcal{P}, \mathcal{R}, \gamma \right\rangle $, where $\mathcal{O}$ is the observation space with a stack of consecutive images $\bm{o}_{t} = \left(\bm{x}_{t-k}, \ldots , \bm{x}_{t-1}, \bm{x}_t\right) $, $\mathcal{A}$ is the action space, $\mathcal{P} \left(\bm{o}_{t+1} | \bm{o}_t, \bm{a}_t\right) $ defines the transition function,  $\mathcal{R}: \mathcal{O} \times \mathcal{A} \rightarrow \mathbb{R} $ denotes the reward function, and $\gamma \in [0,1)$ is the discount factor. Starting from the initial state $\bm{o}_0 \in \mathcal{O}$, the objective is to learn an optimal policy $\pi^*$ that maximizes the expected sum of discount rewards $\mathbb{E}_{\bm{a}_t \sim \pi \left(\cdot | \bm{s}_t\right), \bm{o}_t \sim \ \mathcal{P} }\left[ \sum_{t = 1}^{\infty} \gamma^t r_t  \right]$.

\section{METHOD}\label{method_sec}
\subsection{Visual Reinforcement Learning Backbone}
DrQ-v2 \cite{yarats2022mastering}, an extension of DDPG \cite{lillicrap2015continuous}, is an off-policy RL algorithm for visual continuous control. Clipped double Q-learning \cite{fujimoto2018addressing} and $n$-step returns are integrated into DrQ-v2 to reduce the overestimation bias and better estimate the temporal difference (TD) error, respectively. For each update, a batch of transitions $\mathcal{B} = \left(\bm{o}_t, \bm{a}_t, r_{t:t+n-1}, \bm{o}_{t+n}\right) $ is sampled from the replay buffer $\mathcal{D}$ and then used to derive the critic loss:
\begin{equation}
  \mathcal{L}_Q = \mathbb{E}_{\mathcal{B} \sim \mathcal{D} }\left[ \left( Q_{\phi_k} \left(\bm{h}_t , \bm{a}_t\right) - y  \right)^2  \right] \quad \forall k \in \left\{1,2\right\},
\label{drqv2_critic_loss}
\end{equation}
where $\bm{h}_t = f_{\xi }\left(\text{aug}\left(\bm{o}_t\right)\right)$, $\text{aug}\left(\cdot\right)$ is the customized data augmentation method, and $f_\xi$ is the visual encoder. The $n$-step TD target $y$ is given as 
\begin{equation}
\label{drq_td_target}
  y = \sum_{i = 0}^{n-1} \gamma^i r_{t+i} + \gamma^n \underset{k=1,2}{\rm{min}} Q_{\bar{\phi}_{k}} \left(\bm{h}_{t+n}, \bm{\tilde{a}}_{t+n} \right),
\end{equation}
where 
\begin{equation*}
\label{exploration_noise}
        \bm{\tilde{a}}_{t+n} = \pi_\theta\left( \cdot | \bm{h}_{t+n} \right)+\epsilon,\ \epsilon = \text{clip} \left(\mathcal{N} \left(0, \sigma^2 \right), -c, c \right).
\end{equation*}
The weights of the target critic $\bar{\phi}_{k}$ are updated by Polyak averaging. The policy is learned by minimizing the following loss:
\begin{equation}
  \mathcal{L}_\pi = -\mathbb{E}_{\bm{o}_t \sim \mathcal{B} }\left[ \min_{k=1,2} Q_{\phi_k} \left(\bm{h}_t, \bm{\tilde{a}}_t\right)\right], \ \bm{\tilde{a}}_{t} \sim \pi_\theta\left( \cdot | \bm{h}_{t} \right) + \epsilon.  
\label{drqv2_actor_loss}
\end{equation}
A linear decay schedule or constant can be specified for the standard deviation of the exploration noise according to the task complexity.

\begin{figure}[t]
	\centering
	\includegraphics[width=0.4\textwidth]{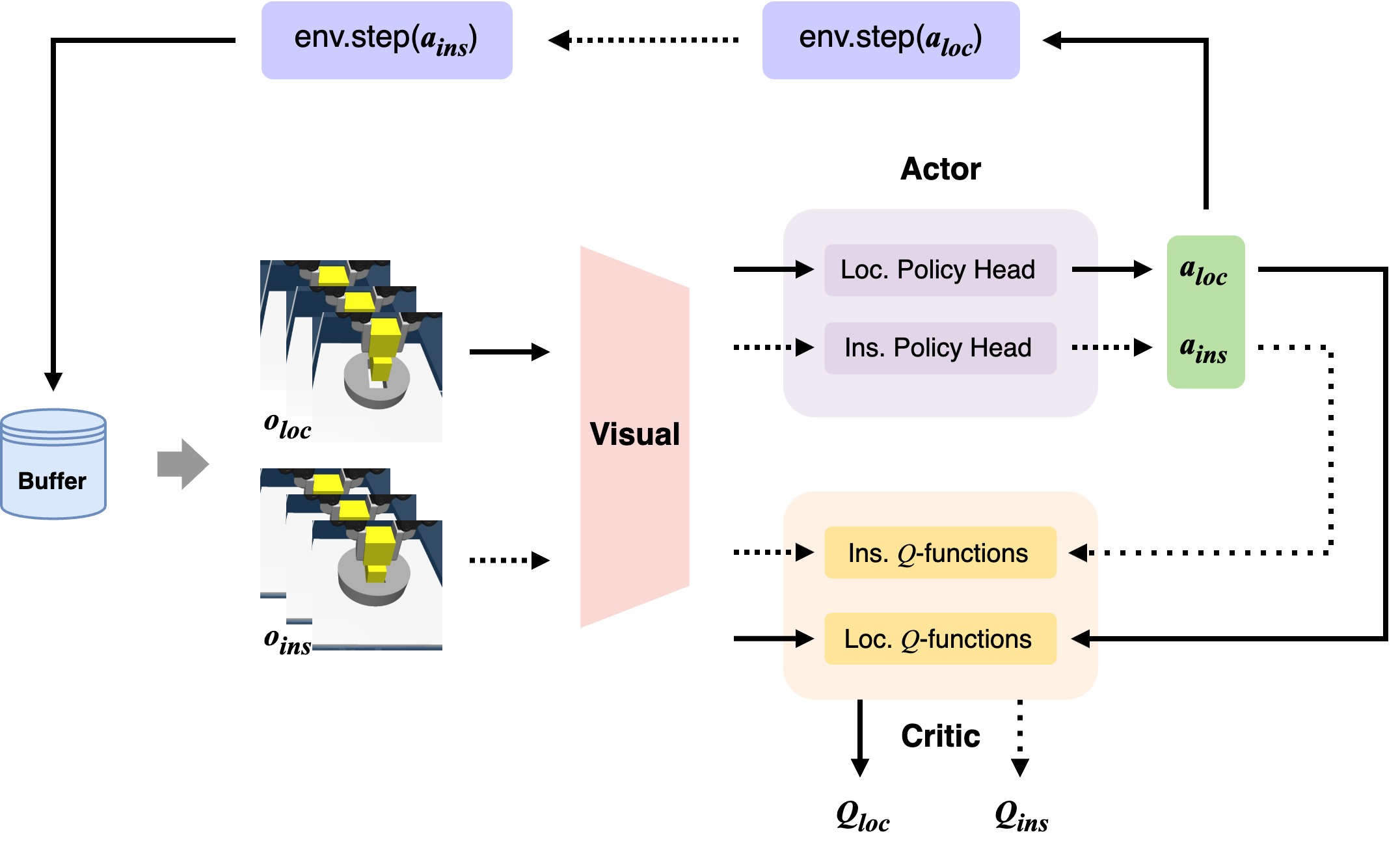}
	\caption{Overview of the proposed insertion strategy. The encoded visual representations are passed to the actor to infer location and insertion actions, which will be performed sequentially. The output actions are fed to the critic with an ensemble of $Q$-functions to evaluate the corresponding $Q$-values.}
	\label{overview}
\end{figure}

\begin{figure}[t]
	\centering
	\includegraphics[width=0.4\textwidth]{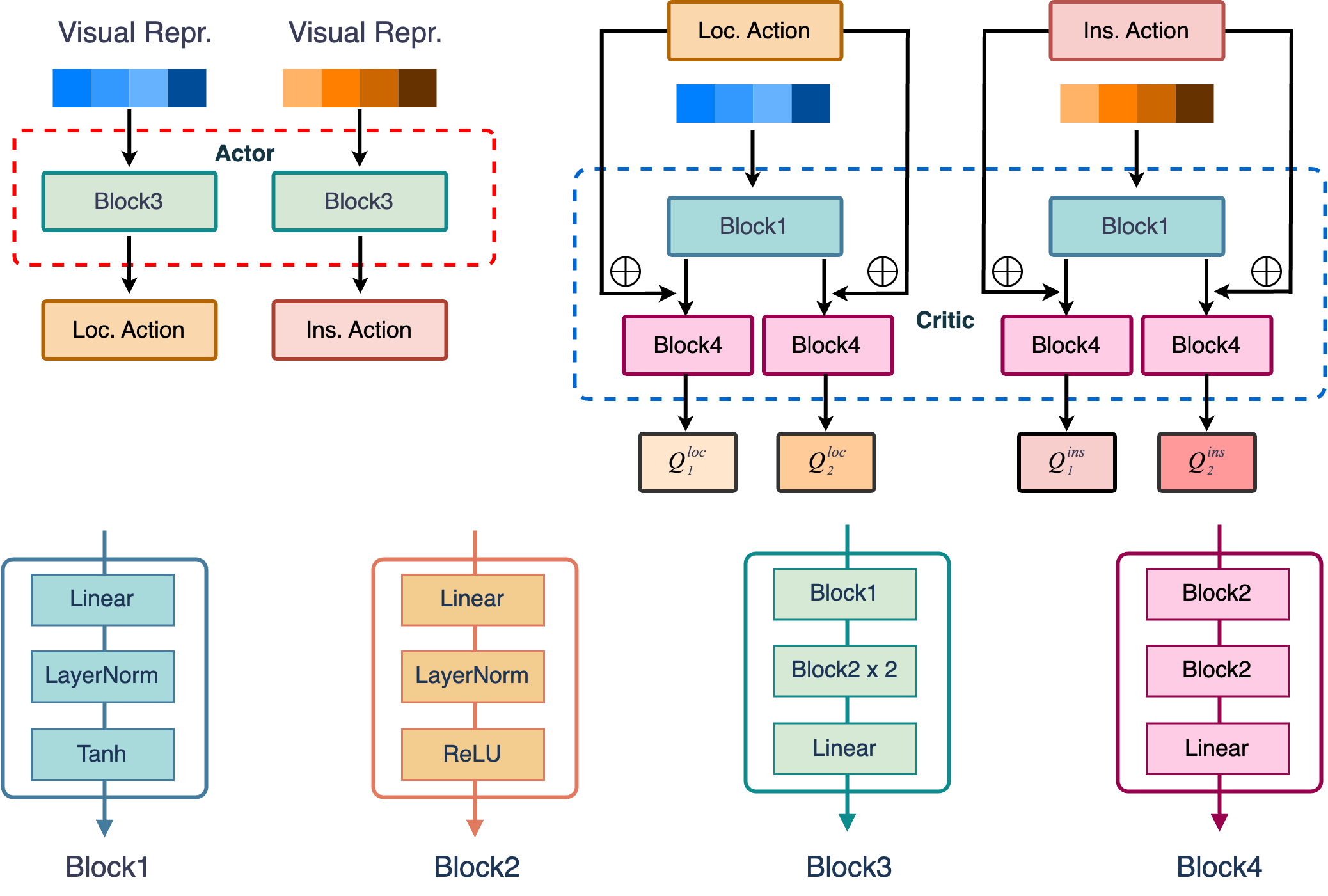}
	\caption{Network architectures for the actor and critic of S2P-DrQ-v2.}
	\label{pipelines}
\end{figure}

\subsection{Separate Primitive Policy}\label{sec: S2P}
DrQ-v2 is treated as the visual RL backbone to exhibit the compatibility of the proposed insertion strategy for the actor-critic framework. Two primitives, location $\bm{a}_{loc}$ and insertion $\bm{a}_{ins}$, are designed to control the predefined behaviors of the agent, which are manifested as the output commands on certain dimensions. The actor integrates two policy networks with the same structure to jointly learn the expressions of the above two actions \cite{huang2020one, shang2023active}. As shown in Fig.~\ref{overview}, location and insertion run sequentially without intricate judgment as to which is chosen at different moments. Each primitive action is deduced separately based on the corresponding observed state, i.e., $\bm{o}_{loc} \rightarrow \bm{a}_{loc}$ and $\bm{o}_{ins} \rightarrow \bm{a}_{ins}$. Thus, POMDP of S2P can be redefined as $\left\langle \mathcal{O}_{loc}, \mathcal{A}_{loc}, \mathcal{P}_{loc}, \mathcal{O}_{ins}, \mathcal{A}_{ins}, \mathcal{P}_{ins}, \mathcal{R}, \gamma \right\rangle $, where $\mathcal{P}_{loc} \left(\bm{o}_{t+1}^{loc} | \bm{o}_t^{loc}, \bm{a}_t^{loc}\right)$ and $\mathcal{P}_{ins} \left(\bm{o}_{t+1}^{ins} | \bm{o}_t^{ins}, \bm{a}_t^{ins}\right)$. Observations $\left(\bm{o}_t^{loc}\ \text{and}\ \bm{o}_t^{ins}\right) $ are stacks of consecutive RGB frames. The step reward saved in the replay buffer is equal to the reward triggered by the insertion primitive. The sample batch is composed of $ \left( \bm{o}_t^{loc}, \bm{a}_t^{loc}, \bm{o}_{t+n}^{loc}, \bm{o}_t^{ins}, \bm{a}_t^{ins}, \bm{o}_{t+n}^{ins}, r_{t:t+n-1}\right) $. An ensemble of four $Q$-functions is employed to construct the double $Q$-learning for each state-action pair, i.e., $Q_{\phi_k} \left(\bm{h}_t^{loc}, \bm{a}_t^{loc} \right)$ and $Q_{\varphi_k}\left(\bm{h}_t^{ins}, \bm{a}_t^{ins} \right), \forall k=1,2$. Unlike performing the same action (e.g., action repeat), the primitives differ from each other to demonstrate distinct behaviors. The definition of primitives is task-specific and presented in Sec.~\ref{evaluation_sec}. $\bm{a}_{loc}$ and $\bm{a}_{ins}$ are executed sequentially within an episode to adjust the relative pose between the peg and hole until success.

\textcolor{black}{	
	\begin{algorithm}[t]
	\begin{algorithmic}[1]
		\caption{Training details for S2P-DrQ-v2}
		\label{alg:algo}
		\Require Actor $\pi_\theta, \pi_\vartheta$, Critic $Q_{\phi_k}, Q_{\varphi_k}$, visual encoder $f_\xi$, learning rate $\eta$, replay buffer $\mathcal{D}$, training frames $T$, exploration noise schedule $\sigma_t^{loc}, \sigma_t^{ins}$, standard deviation clip intervals $c_{loc}, c_{ins}$, batch size, Polyak average rate $\mu$, and UTD ratio.
		\For {$t = 1 \ldots T$}
			\State $\bm{a}_t^{loc} \leftarrow \pi_\theta \left(\cdot | f_\xi \left(\text{aug} \left(\bm{o}_t^{loc}\right)\right) \right) $
			\State $\bm{o}_{t+1}^{loc} \leftarrow \mathcal{P}\left(\cdot | \bm{o}_t^{loc}, \bm{a}_t^{loc} \right)$
			\State $\bm{a}_t^{ins} \leftarrow \pi_\vartheta \left(\cdot | f_\xi \left(\text{aug} \left(\bm{o}_t^{ins}\right)\right)  \right)$
			\State $\bm{o}_{t+1}^{ins} \leftarrow \mathcal{P}\left(\cdot | \bm{o}_t^{ins}, \bm{a}_t^{ins} \right)$
			\State $\mathcal{D}  \leftarrow \mathcal{D} \cup \left( \bm{o}_t^{loc}, \bm{o}_t^{ins}, \bm{a}_t^{loc}, \bm{a}_t^{ins}, r_t, \bm{o}_{t+1}^{loc}, \bm{o}_{t+1}^{ins} \right) $
			\State $\mathcal{B} \leftarrow \mathcal{D}$ \hfill $\qquad \vartriangleright \text{Sample batch transitions}$
			\For {$j=1 \ldots \text{UTD}$}
				\State \textsc{UpdateCritic}$\left(\mathcal{B}, \sigma_t^{loc}, \sigma_t^{ins}\right)$ with Eq.~\ref{new_drqv2_critic_loss}
				\State \textsc{UpdateActor}$\left(\mathcal{B}, \sigma_t^{loc}, \sigma_t^{ins} \right)$ with Eq.~\ref{new_drqv2_actor_loss}
			\EndFor
		\EndFor
	\end{algorithmic}
\end{algorithm}
}
The original network architecture of DrQ-v2 is modified to support S2P, which is shown in Fig.~\ref{pipelines}. Thus, Eqs.~\ref{drqv2_critic_loss}-\ref{drqv2_actor_loss} are reformulated on the basis of the above statement:
\begin{align}
	y &= \sum_{i = 0}^{n-1} \gamma^i r_{t+i} + \gamma^n  \bigl( \underset{k=1,2}{\rm{min}} Q_{\bar{\phi}_{k}} \left(\bm{h}_{t+n}^{loc}, \bm{\tilde{a}}_{t+n}^{loc} \right) \nonumber \\
	& \hspace{2.cm} + \underset{k=1,2}{\rm{min}} Q_{\bar{\varphi}_{k}} \left(\bm{h}_{t+n}^{ins}, \bm{\tilde{a}}_{t+n}^{ins} \right) \bigr) \label{new_drq_td_target} \\
	\mathcal{L}_Q & = \mathbb{E}_{\mathcal{B} \sim \mathcal{D} } \biggl[ \biggl( \bigl( Q_{\phi_k} \left(\bm{h}_t^{loc} , \bm{a}_t^{loc}\right) \nonumber \\
	& \hspace{1.5cm} + Q_{\varphi_k} \left(\bm{h}_t^{ins}, \bm{a}_t^{ins} \right) \bigr)  - y  \biggr)^2  \biggr]  \label{new_drqv2_critic_loss} \\
	\mathcal{L}_{\pi} &= -\mathbb{E}_{\bm{o}_t^{loc}, \bm{o}_t^{ins} \sim \mathcal{B} } \biggl[ \min_{k=1,2} Q_{\phi_k} \left(\bm{h}_t^{loc}, \bm{\tilde{a}}_t^{loc} \right) \nonumber \\
	& \hspace{2.5cm} + \min_{k=1,2} Q_{\varphi_k} \left(\bm{h}_t^{ins}, \bm{\tilde{a}}_t^{ins}\right) \biggr], \label{new_drqv2_actor_loss} 
\end{align}
where $\bm{a}^{loc}$ and $\bm{a}^{ins}$ are sampled from the location $\pi_\theta$ and insertion $\pi_\vartheta$ policies, respectively. Exploration noise for different actions follows the same linear schedule. However, different noise clip intervals (i.e., $\epsilon_{loc} $ and $ \epsilon_{ins}$) can be specified to achieve desired explorations:
\begin{align}
	\bm{\tilde{a}}_{t}^{loc} & \sim \pi_\theta\left( \cdot | \bm{h}_{t}^{loc} \right) + \epsilon_{loc} \nonumber \\
	\bm{\tilde{a}}_{t}^{ins} & \sim \pi_\vartheta \left( \cdot | \bm{h}_{t}^{ins} \right) + \epsilon_{ins}, 
	\label{clip_intervals} 
\end{align}
where $\epsilon_{loc} \sim \text{clip} \left(\mathcal{N} \left(0, \sigma_{loc}^2\right), -c_{loc}, c_{loc} \right) $ and $\epsilon_{ins} \sim \text{clip} \left(\mathcal{N} \left(0, \sigma_{ins}^2\right), -c_{ins}, c_{ins} \right)$. The pseudocode in Algorithm~\ref{alg:algo} illustrates the interference and training details of S2P-DrQ-v2.

\begin{figure*}[t]
	\centering
	\includegraphics[width=0.6\textwidth]{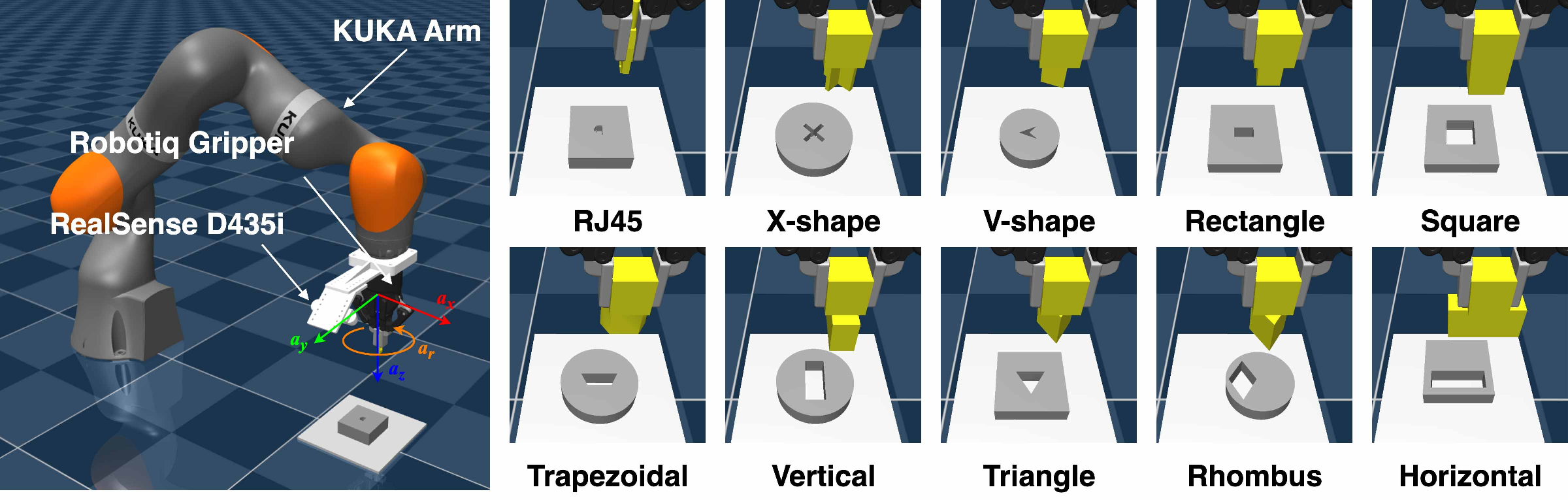}
	\caption{Simulation setup and peg-in-hole suites with different shapes, where pegs are initialized with being grasped by the gripper.}
	\label{setup}
\end{figure*}

In contrast to the original DrQ-v2, S2P can improve the sample efficiency. Generally, the collected samples need to cover as many state-action pairs as possible:
\begin{equation}
	N_s^{org} \propto \left\lvert \mathcal{O} \right\rvert \times k^{D},
\end{equation}
where $\mathcal{O}$ denotes the observation space, $k$ represents the required samples for each action dimension, and $D$ indicates the dimension of the action space. To accomplish peg-in-hole tasks, the required samples should theoretically cover generalized location $\bm{a}_{loc} \in \mathbb{R}^{D_{loc}}$ and insertion $\bm{a}_{ins} \in \mathbb{R}^{D_{ins}}$ actions
\begin{equation}
	N_s^{org} \propto \left\lvert \mathcal{O} \right\rvert \times k^{D_{loc}+{D_{ins}}}.
\end{equation}
However, S2P reconstructs the original action space. The number of required samples for S2P is reduced to
\begin{equation}
	N_s^\mathit{S2P} \propto \left\lvert \mathcal{O} \right\rvert \times \left( k^{D_{loc}}+k^{D_{ins}}\right).  
\end{equation}
The boosted sample efficiency can be estimated as
\begin{equation}
	\frac{N_s^{org}}{N_s^\mathit{S2P}} = {\frac{k^{D_{loc}+{D_{ins}}}}{k^{D_{loc}}+k^{D_{ins}}}}.  
\end{equation}
Besides, the critic networks can be regarded as a nonlinear approximator to assign $Q$-values to state-action pairs. A large action space will increase the estimation bias \cite{zhu2021overview}. S2P decomposes the original $Q$-value with the following structural prior
\begin{equation}
	Q \left(\bm{h}_t, \bm{a}_t\right) = Q_{\phi} \left(\bm{h}_t^{loc}, \bm{a}_t^{loc}\right) + Q_{\varphi} \left(\bm{h}_t^{ins}, \bm{a}_t^{ins}\right).  
\end{equation}
For some complex scenarios, the same observations with slight action variations can result in quite different rewards. More accurate $Q$-value estimation prevents the policy from falling into suboptimality and exploration dilemmas. The update of the actor in S2P-DrQ-v2 depends on decoupled gradients
\begin{align}
	\nabla_\theta J(\theta) \approx &\ \mathbb{E}\left[ \nabla_\theta \log \pi_\theta(\cdot | \bm{h}_t^{loc}) \cdot Q_{\phi}(\bm{h}_t^{loc}, \bm{a}_t^{loc}) \right] \nonumber \\
	\nabla_\vartheta J(\vartheta) \approx &\ \mathbb{E}\left[ \nabla_\vartheta \log \pi_\vartheta(\cdot | \bm{h}_t^{ins}) \cdot Q_{\varphi}(\bm{h}_t^{ins}, \bm{a}_t^{ins}) \right]
\end{align}
with clearer signal. Thus, S2P decomposes a high-dimensional, complex joint optimization problem into two low-dimensional, simpler subproblems.

\begin{figure*}[t]
	\centering
	\includegraphics[width=0.65\textwidth]{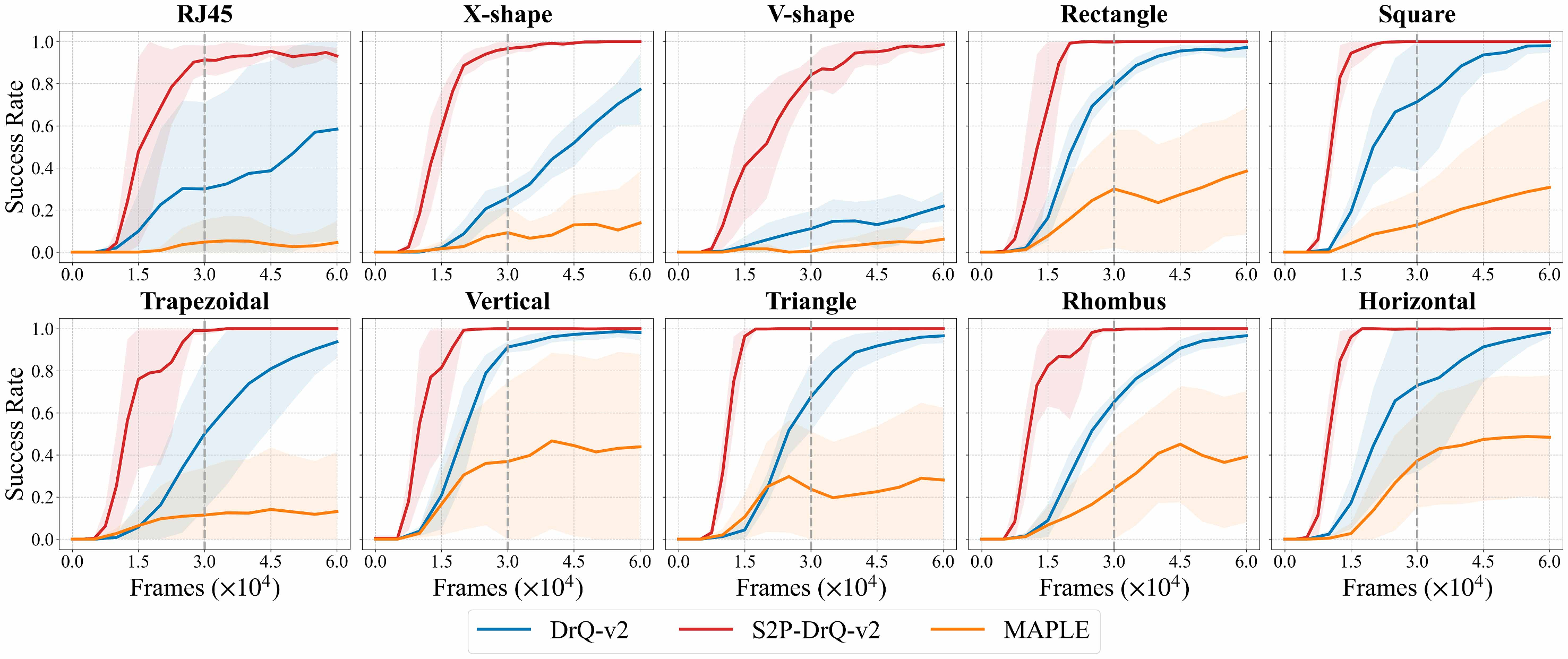}
	\caption{Training performance of S2P against baselines, where the solid line and the shaded area represent the mean and standard deviation across 5 runs.}
	\label{rr_2}
\end{figure*}

\section{EXPERIMENT}
\subsection{Setup}\label{setup_sec}
All tests in this paper are carried out on the platform shown in Fig.~\ref{setup}, where a connector is utilized to attach the Robotiq 2F-85 gripper to the KUKA LBR IIWA 14 arm. RealSense D435i is fixed on the camera mount, and only RGB images are collected both in simulation and real-world experiments. The setup in Fig.~\ref{setup} is intended to match the real-world scenario. Meanwhile, the operational space controller \cite{khatib1987unified} is deployed to enable the compliance contact:
\begin{equation}
	\bm{\tau}_{op} =\bm{J}^T \bm{\Lambda} \bigl(- \bm{K}_d \left(\bm{\dot{x}}- v \dot{\bm{x}}_d\right)  \bigr) + \bm{C} + \bm{G},
\end{equation}
where $\bm{J} \in \mathbb{R}^{6\times 7} $ is the task Jacobian matrix. $\bm{\Lambda} \in \mathbb{R}^{6 \times 6}$ is the operational space inertial matrix. $\dot{\bm{x}}_d = {\bm{K}_p}/{\bm{K}_d}\left(\bm{x}_d - \bm{x}\right)$ and $\bm{x}_d$ are the desired velocity and position of the end-effector, respectively. $\bm{K}_p, \bm{K}_d \in \mathbb{R}^{6} $ are diagonal elements of PD gains. $\bm{C} \in \mathbb{R}^7 $ is the Coriolis and Centrifugal forces, and $\bm{G} \in \mathbb{R}^7$ is the gravity vector. $v$ is the saturation function to limit the maximum end-effector velocity $\bm{v}_{max}$, which is defined as
\begin{equation}
	v=\min\left(1, {\bm{v}_{max}} / {\sqrt{\bm{\dot{x}}_d^T \bm{\dot{x}}_d}} \right).   
\end{equation}
Further stabilization control of the redundant manipulator is enabled with the nullspace filter
\begin{equation}
	\bm{\tau}_{null} = \left(\bm{I} - \bm{J}^T \bar{\bm{J}}^T\right)  \left(-\bm{K}_{qd}\bm{M}\dot{\bm{q}} \right),
	\label{nullspace}   
\end{equation}
where $\bar{\bm{J}}=\bm{M}^{-1}{\bm{J}}^T\bm{\Lambda}$ is the generalized inverse of the Jacobian matrix. $\bm{M} \in \mathbb{R}^{7\times7}$ is the manipulator inertial matrix. $\bm{K}_{qd} \in \mathbb{R}^7$ is the derivative gain in the joint space. $\dot{\bm{q}} \in \mathbb{R}^7$ is the joint velocity vector. The joint torque command is 
\begin{equation}
	\bm{\tau} = \bm{\tau}_{op} + \bm{\tau}_{null}.
	\label{torque}   
\end{equation}
Fig.~\ref{setup} depicts that the agent is endowed with 4 feasible DOFs, i.e., $\left(a_x,a_y,a_z,a_r\right) $, including the movement in Cartesian space and the rotation around the yaw axis. Each training episode is reset with a randomized end-effector pose within $\pm [0.02, 0.02, 0.03]$ m and $ \pm 0.3$ rad intervals. We integrate a suite of peg-in-hole tasks containing different types of pegs and holes, which are shown in Fig.~\ref{setup}. Each task is initialized with the peg already grasped by the gripper. This suite, powered by MuJoCo, contains different pegs and holes with basic polygons and extra challenging shapes. RJ45, X-shape, and V-shape are freshly designed, and the rest are inherited from the open-sourced assets\footnote{\url{https://github.com/carlosferrazza/tactile_envs.git}}.

\subsection{Evaluations}\label{evaluation_sec}
\textbf{Training Details}. Three consecutive frames with $96 \times 96$ size are adopted as observations that are passed to the visual encoder to obtain the low-dimensional representation. Random shift \cite{yarats2022mastering} is employed to obtain augmented data. 60K training frames are assigned to all tasks. Success rate is utilized as the evaluation metric to observe the sample efficiency and performance. A staged dense reward that applies to all tasks is designed to facilitate convergence:
\begin{equation}
	r_t = \max \left\{
	\begin{alignedat}{3}
		& w_r \left(1 - \tanh\left(\lambda \cdot \triangle s_1 \right) \right) (\text{reaching}) \\
		& 3 + w_a \left(1- \triangle s_1 / \varepsilon _a \right) \\
		& \quad \quad \text{if } \triangle s_1 \leq \varepsilon _a \ \& \ \triangle \delta \leq \delta _a (\text{alignment})  \\ 
		& 4 + w_i h \quad \\
		& \quad \quad \text{if} \ h \leq 0 \ \& \ \text{alignment achieved}\  (\text{insertion}) \\
		& 10 \quad \text{if} \ \triangle s_2 \leq \varepsilon _f \ (\text{completion}), \\
	\end{alignedat}
	\right.
	\label{reward}
\end{equation}
where $w_r$, $w_a$, and $w_i$ are reward weights for different stages. The relative distances from the peg tip to the surface and bottom of the hole are indicated as $\triangle s_1$ and $\triangle s_2$, respectively. $\triangle \delta$ is the orientation error. $\varepsilon _a$, $\delta _a$, and $\varepsilon _f$ are pose thresholds at different stages. $h$ is the insertion depth. Once the completion signal is detected, the current episode will be marked as successful and then truncated. The whole training process is accomplished with an NVIDIA 4080 GPU and an i7-12700K CPU. The comparison results are obtained using the following methods:
\begin{itemize}
\item S2P-DrQ-v2: Two separate policies are integrated in the actor of DrQ-v2, which deduces the location $\bm{a}_{loc} \coloneqq \left(a_x, a_y, a_z, a_r\right) $ and insertion $\bm{a}_{ins} \coloneqq \left(0, 0, a_z, 0\right)$ primitives, respectively.
\item DrQ-v2: A basic implementation only derives $\bm{a} \coloneqq \left(a_x, a_y, a_z, a_r\right)$.
\item MAPLE \cite{nasiriany2022augmenting}: A representative of hierarchical RL consists of a high-level categorical policy and a low-level primitive policy. MAPLE is optimized for the visual input, and its primitive definitions are the same as those of S2P-DrQ-v2.
\end{itemize}

As described in Sec.~\ref{method_sec}, S2P-DrQ-v2 sequentially executes location and insertion primitives that consume two environment steps. In contrast, action repeat (AR) only implies the execution of the same action in consecutive steps. To align the consistent maximum episode length with S2P-DrQ-v2, AR = $2$ is assigned to DrQ-v2. All the above methods are trained on the premise of $\text{UTD} = 2$ to improve the underlying sample efficiency. The hyperparameters used for training are given in Tables \ref{table: common_hyper} and \ref{table: special_hyper} of the Appendix.

\textbf{Evaluation Results}. Fig.~\ref{rr_2} presents the training results for ten tasks, where each policy is evaluated through 5 different seeds. S2P-DrQ-v2 outperforms DrQ-v2 in terms of sample efficiency. Meanwhile, the exploration of location policy can be narrowed through $\bm{a}_{ins}$ to output deterministic location actions. $\bm{a}_{ins}$ is scheduled in advance as a chain of sequential actions to increase the number of insertion attempts. Although AR can be used to enable similar exploration and reinforce perception of the current action, we found this is less effective than establishing a policy to learn the insertion action individually. As shown in Fig.~\ref{rr_2}, for the challenging V- and X-shape tasks, S2P is able to substantially improve the success rate while enhancing the sample efficiency. The gray dashed lines in Fig.~\ref{rr_2} indicate the success rates of different methods in the middle of the training period to quantify the performance comparison. For most tasks, S2P-DrQ-v2 achieves a performance gain of 20\%-30\% relative to DrQ-v2, especially for the case with complex polygons. Notably, S2P-DrQ-v2 attains a $\sim100\%$ success rate with only half of the training frames. MAPLE is built upon Soft Actor-Critic (SAC) \cite{haarnoja2018soft}. Yarats et al. \cite{yarats2022mastering} have shown that DrQ-v2 achieves better sample efficiency than SAC. For MAPLE, two different temperature coefficients are used to balance the exploration and exploitation of high- and low-level policies, which exacerbates the low sample efficiency of MAPLE. Numerous primitives are not needed for the peg-in-hole task. Thus, co-training with a high-level categorical policy to select and combine different primitives increases the training cost.

\begin{figure}[t]
	\centering
	\includegraphics[scale=0.18]{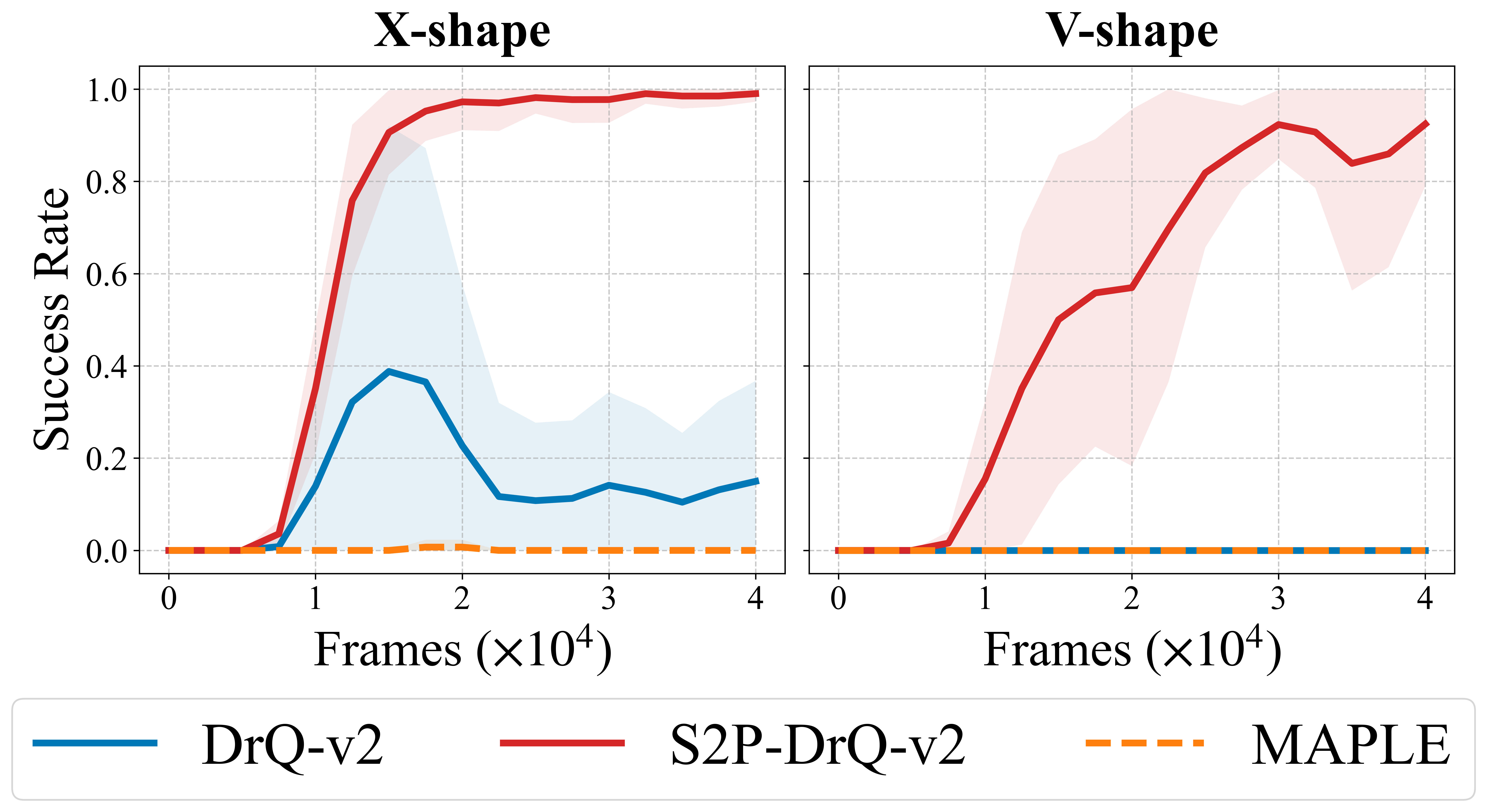}
	\caption{Success rates of S2P and baselines under force penalty.}
	\label{rr_force}
\end{figure}

For contact-rich tasks, large contact forces generated during the exploration phase can lead to hardware wear and potential safety risks. Furthermore, we attempt to evaluate the ability of S2P to find the optimum state with multiple reward and penalty conditions. Thus, a force threshold of 15 N is given to prevent the agent from learning aggressive behaviors, i.e., 
\begin{equation}
	r_{f} = -w_f \cdot \mathds{1} \big[\left\lvert f_{ee}\right\rvert  > \left\lvert f_{max}\right\rvert \big] \cdot \Vert f_{ee} - f_{max} \Vert_2,
\end{equation}
where $\mathds{1}\left[\cdot\right] $ is the indicator function. The end-effector load is measured first in the simulation environment, and thus accurate external force data can be obtained from the sensor. Only representative X- and V-shape tasks are adopted for comparison to save computational resources. The location policy of S2P and MAPLE derives $\bm{a}_{loc}$ without movement along the $z$-axis in Cartesian space, allowing the insertion policy to specialize in learning compliant insertion behavior. Thus, $\bm{a}_{loc}$ is defined as $\left(a_x,a_y,0,a_r\right)$ under force constraints and as $\left(a_x,a_y,a_z,a_r\right)$ in unconstrained scenarios. Meanwhile, the action sequence of DrQ-v2 is planned as $\bm{a}_{loc} = \left(a_x, a_y, 0, a_r\right) \rightarrow \bm{a}_{ins} = \left(0, 0, a_z, 0\right)$ to mimic the locate-then-insert behavior of S2P. Fig.~\ref{rr_force} indicates that both DrQ-v2 and MAPLE fail to complete the aforementioned tasks. The entire training process of DrQ-v2 is unstable and non-convergent. The additional force penalty has a slight impact on the sample efficiency of S2P, rivaling the benchmark results in Fig.~\ref{rr_2}. These comparison results further demonstrate that learning the insertion primitive via a separate policy enables more flexible adaptation to diverse scenarios.

\subsection{Real-world Experiments}
The feasibility of S2P is verified in this section by transferring the trained policy to the real platform in a zero-shot manner. To improve the sim2real performance, random overlay \cite{hansen2021generalization} is adopted for acquiring the strong augmented data stream, and the location and insertion primitives are set as $\left(a_x, a_y, a_z\right) $ and $(0,0,a_z)$, respectively. The RL framework implemented in Sec.~\ref{sec: S2P} is replaced with SVEA \cite{hansen2021stabilizing}, which stabilizes the integration of the two different data streams during training. The visual encoder and actor are frozen during the sim2real transfer. Meanwhile, a curriculum-based domain randomization scheme \cite{yuan2024learning} is implemented, including light, appearance, camera pose, and dynamic parameters. The bracket used to hold the wrist camera is first imported into the simulation and then 3D printed in reality. Thus, the range for randomizing the camera positions can be narrowed to reduce training costs.

\begin{figure}[t]
	\centering
	\includegraphics[width=0.38\textwidth]{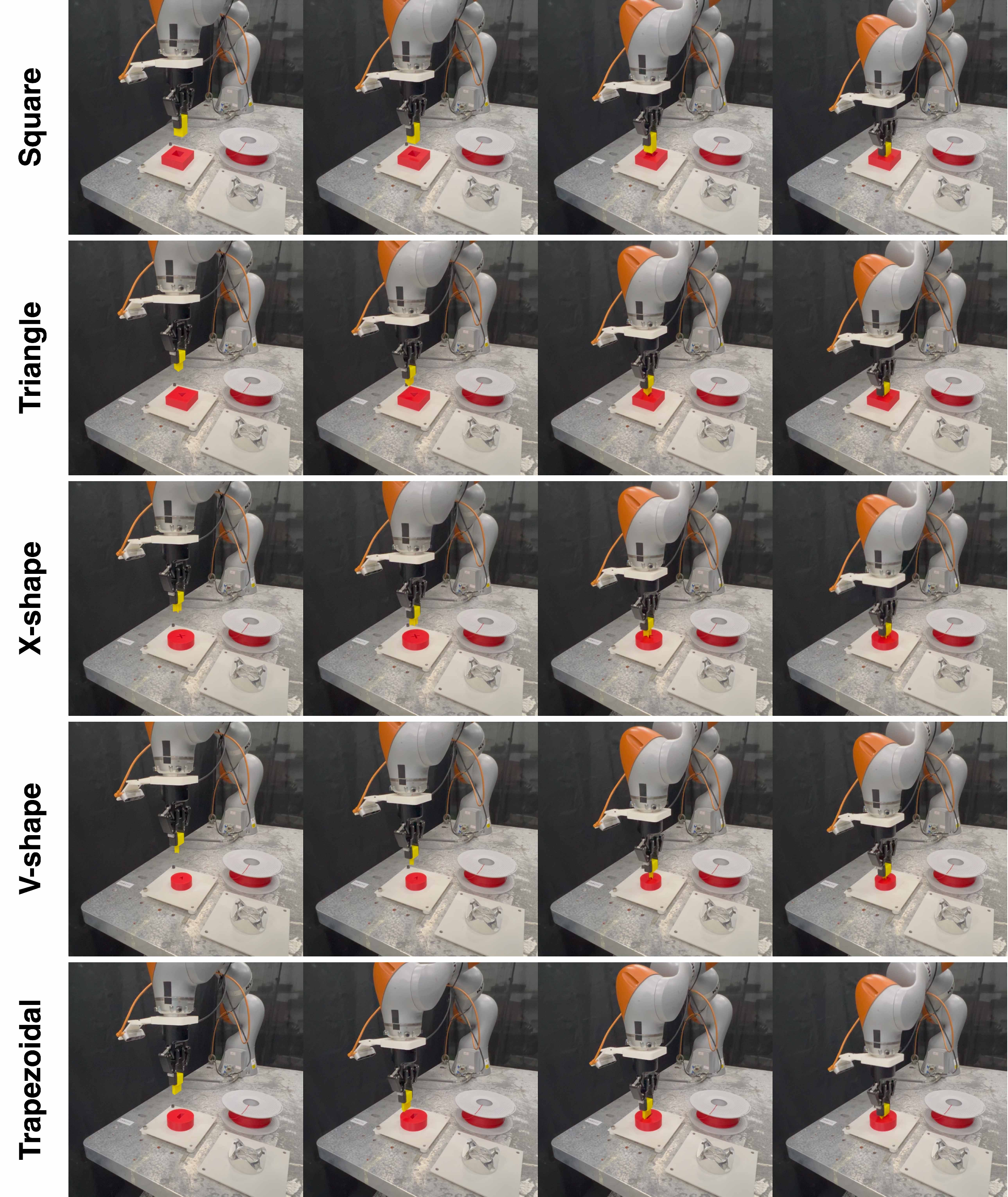}
	\caption{Zero-shot sim2real transfer processes with clutters within the camera viewpoint.}
	\label{exp_process}
\end{figure}

\begin{figure}[t]
	\centering
	\includegraphics[width=0.38\textwidth]{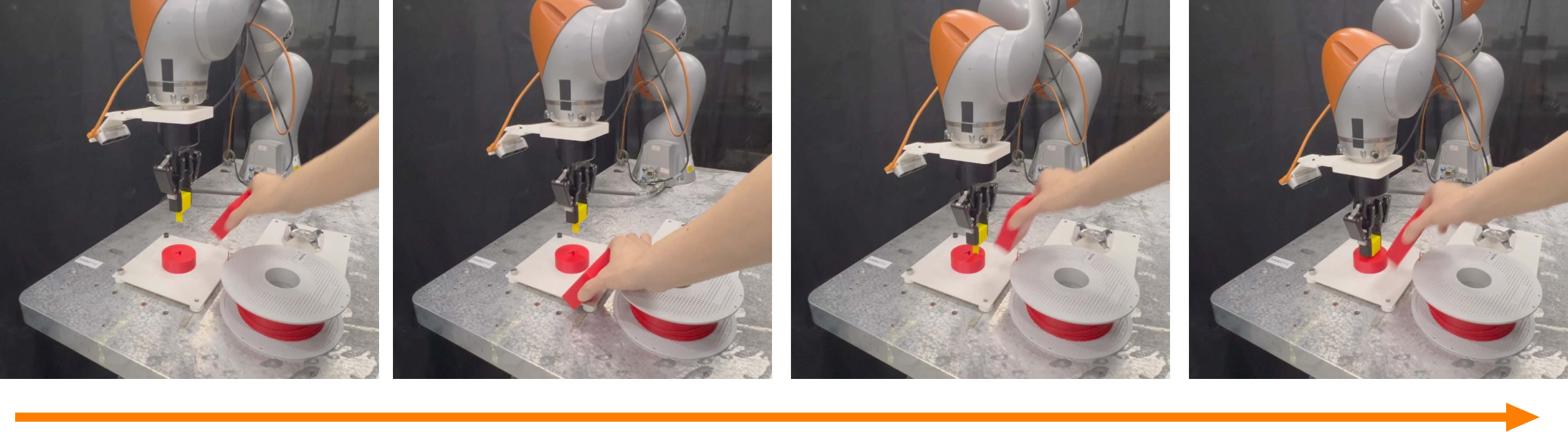}
	\caption{Dynamic visual perturbation during the insertion process.}
	\label{visual_perturbation}
\end{figure}

Five representative tasks are utilized for evaluations, and the setup of the experimental platform is presented in Fig.~\ref{exp_process}. The KUKA controller, RealSense D435i, PC client, and server are the major components of the communication network. The PC client transmits the actor command to the PC server, which constructs the communication bridge between the KUKA controller and the PC client. All data transmission is implemented through socket connections. The Cartesian impedance mode is predefined in the controller and executed in the background. To demonstrate the practicality of visual input, some task-irrelevant distractors are placed around the hole as visual interference, and the hole is placed on the table without additional calibration to align with that in the simulation. Fig.~\ref{exp_process} illustrates that S2P-SVEA exhibits the robustness to visual interference and can be transferred to the real platform in a zero-shot manner. Meanwhile, S2P-SVEA can endure the dynamic visual perturbation in Fig.~\ref{visual_perturbation}. The quantitative experimental results on different tasks are illustrated in Tab.~\ref{test_result}.

\begin{table}[t!]
	\centering
	\caption{Quantitative results of sim2real transfer.}
	\label{test_result}
	\begin{tabular}{@{}>{\centering\arraybackslash}m{1.cm}ccccc@{}}
	\toprule  
	  & Square & Triangle & X-shape & V-shape & Trapezoidal  \\ 
	\midrule
	Succ. Rate & 40 / 40  & 40 / 40 & 40 / 40 & 39 / 40 & 40 / 40 \\
	\bottomrule   
	\end{tabular}
\end{table}

\subsection{Ablations}
This section performs some ablation experiments to discuss the hyperparameters of S2P-DrQ-v2 that could impact the performance and the generalization for other off-policy algorithms like SAC. Similarly, for computational resources, the following ablation experiments are restricted to force-penalized X- and V-shape tasks.

\begin{figure}[t]
	\centering
	\includegraphics[width=0.35\textwidth]{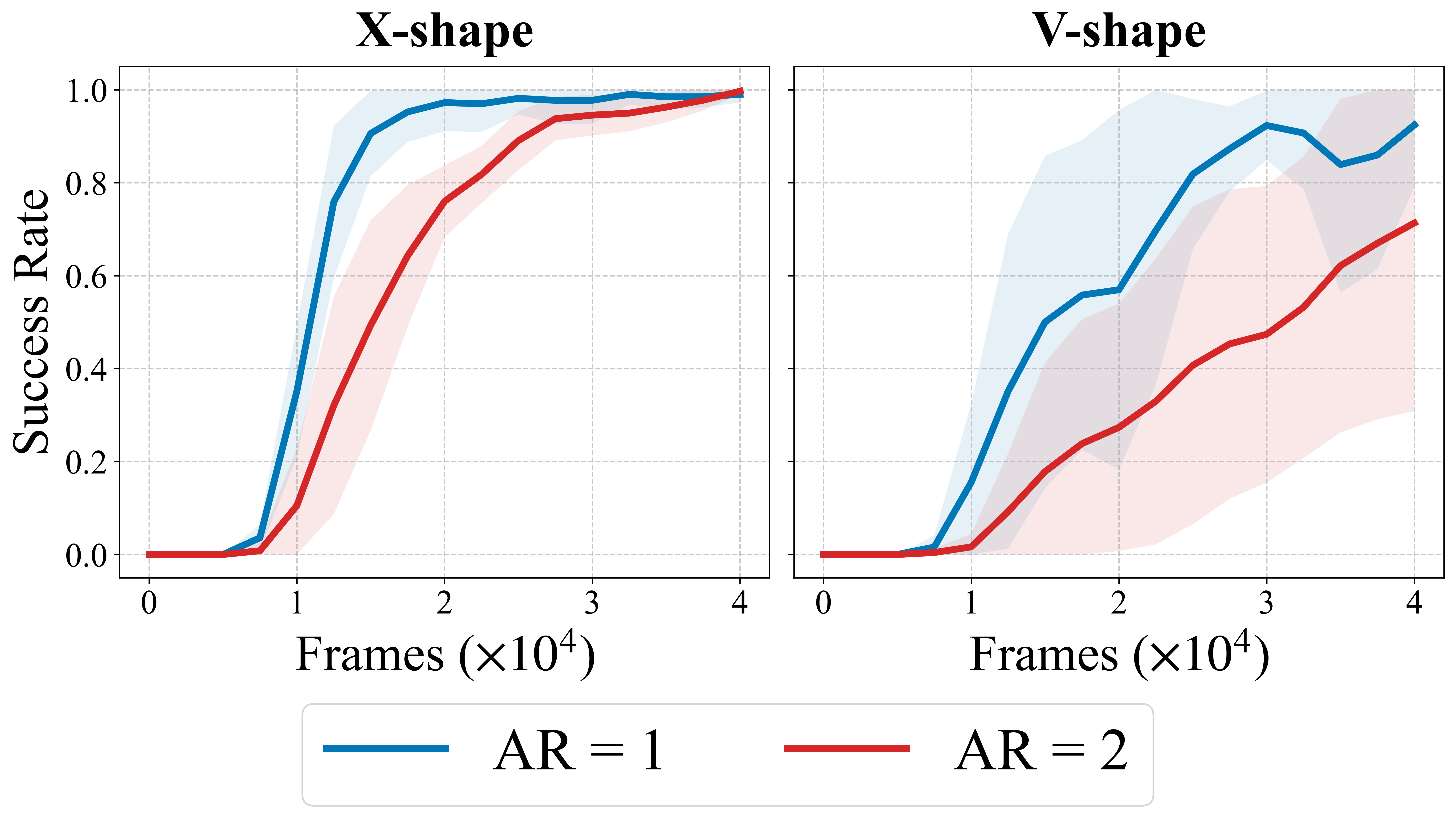}
	\caption{Ablation analysis on the effect of action repeat on S2P-DrQ-v2.}
	\label{action_repeat}
\end{figure}

\textbf{Action repeat.} AR is a widely used technique to enhance the recognition of actions that trigger high rewards and thus accelerate the sample efficiency. S2P requires two consecutive environment steps, which exhibit identical temporal characteristics to $\text{AR}=2$ for DrQ-v2. Therefore, all evaluations on S2P in previous sections are performed without AR. This part will explore whether AR also benefits S2P. $\text{AR}=2$ for S2P indicates that $\bm{a}_{loc}$ and $\bm{a}_{ins}$ are performed sequentially twice and extended to four environment steps. As shown in Fig.~\ref{action_repeat}, $\text{AR}=2$ fails to improve the performance of S2P and decreases the sample efficiency. Especially for the V-shape task, $\text{AR}=2$ oscillates training processes with wider standard deviation intervals. Unlike DrQ-v2, S2P specializes in separate adjustments of location and insertion actions based on corresponding observations, which enables a flexible response to frequent changes in contact states and thus refines primitive actions. Such a property of S2P is efficient when there exist force constraints. However, action repeat extends the reaction period to two steps, which adversely slows the learning process.

\begin{figure}[t]
	\centering
	\includegraphics[width=0.38\textwidth]{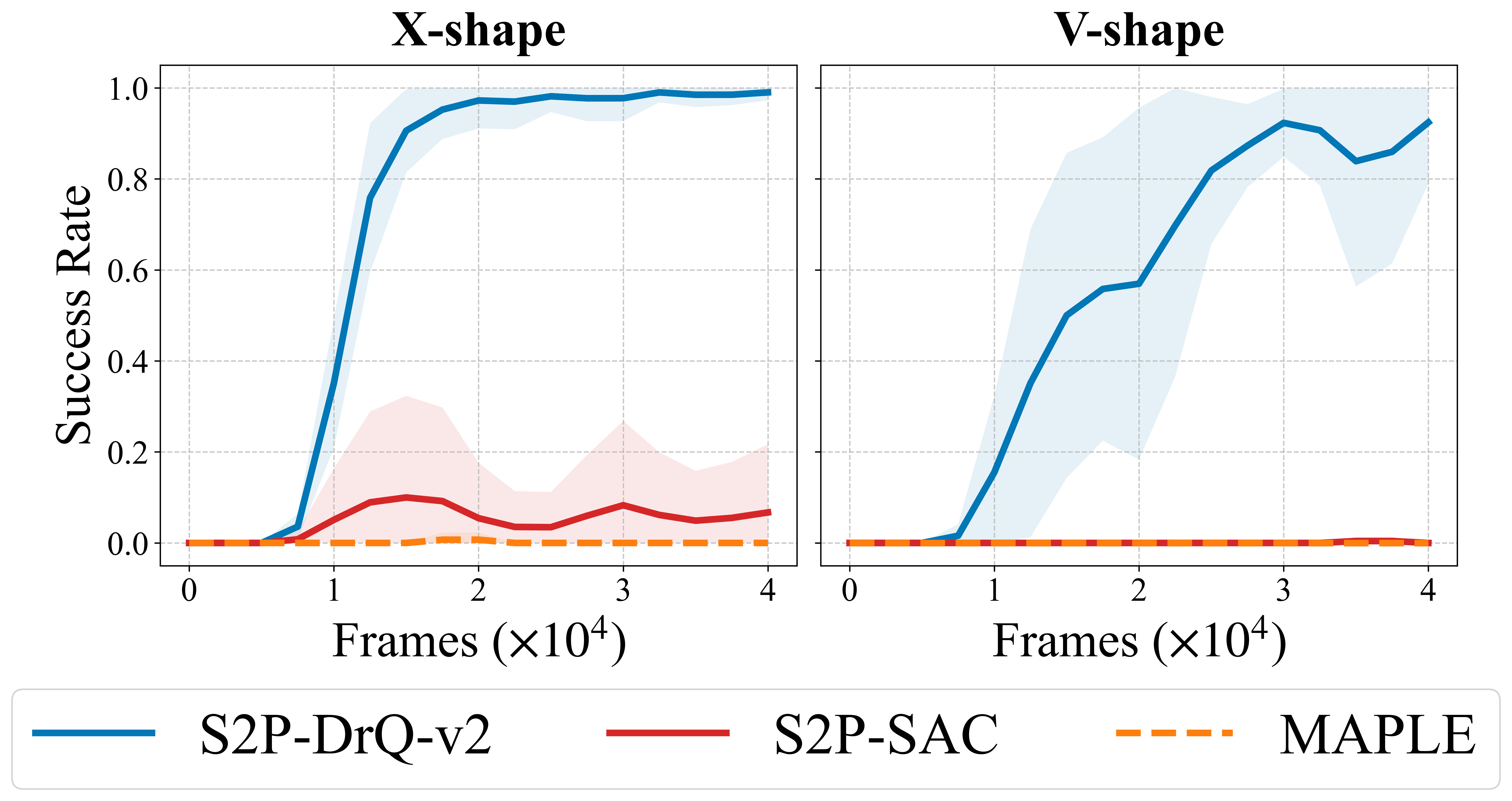}
	\caption{Ablation study on extending S2P to SAC.}
	\label{rr_sac}
\end{figure}

\textbf{Generalization to SAC.} S2P is adapted to the actor-critic framework. Thus, we extend S2P to SAC (i.e., S2P-SAC) to investigate whether similar performance gains can be achieved. The implementation details of S2P-SAC can be summarized as follows:
\begin{align}
	y = & \sum_{i = 0}^{n-1} \gamma^i r_{t+i} + \gamma^n \bigg( \bigl( \underset{k=1,2}{\rm{min}} Q_{\bar{\phi}_{k}} \left(\bm{h}_{t+n}^{loc}, \bm{\tilde{a}}_{t+n}^{loc} \right) - \alpha_{loc} \nonumber \\
	& \log \pi_\theta\left( \bm{\tilde{a}}_{t+n}^{loc} | \bm{h}_{t+n}^{loc} \right) \bigr) + \big( \underset{k=1,2}{\rm{min}} Q_{\bar{\varphi}_{k}} \left(\bm{h}_{t+n}^{ins}, \bm{\tilde{a}}_{t+n}^{ins} \right) - \nonumber \\
	& \alpha_{ins} \log \pi_\vartheta \left( \bm{\tilde{a}}_{t+n}^{ins} | \bm{h}_{t+n}^{ins} \right) \big) \bigg) \\
	\mathcal{L}_\pi = &\ \mathbb{E}_{\bm{o}_t^{loc}, \bm{o}_t^{ins} \sim \mathcal{B}} \bigg[ \bigg(\alpha_{loc} \log \pi_{\theta} \left( \bm{\tilde{a}}_t^{loc} | \bm{h}_t^{loc} \right) - \min_{k=1,2} \nonumber \\
	& Q_{\phi_k} \left(\bm{h}_t^{loc}, \bm{\tilde{a}}_t^{loc} \right)\bigg) + \bigg(\alpha_{ins} \log \pi_{\vartheta} \left( \bm{\tilde{a}}_t^{ins} | \bm{h}_t^{ins} \right) \nonumber  \\
	& - \min_{k=1,2} Q_{\varphi_k} \left(\bm{h}_t^{ins}, \bm{\tilde{a}}_t^{ins} \right) \bigg) \bigg] \\
	\mathcal{L}_{\alpha}^{loc}  =&\ \mathbb{E}_{\bm{\tilde{a}}_t^{loc} \sim \pi_\theta, \bm{o}_t^{loc} \sim \mathcal{B}} \bigg[ - \alpha_{loc} \log \pi_\theta \left( \bm{\tilde{a}}_t^{loc} | \bm{h}_t^{loc} \right)  - \nonumber \\
	& \alpha_{loc} \bar{\mathcal{H}}_{loc}  \bigg] \\
	\mathcal{L}_{\alpha}^{ins} =&\  \mathbb{E}_{\bm{\tilde{a}}_t^{ins} \sim \pi_\vartheta, \bm{o}_t^{ins} \sim \mathcal{B}} \bigg[- \alpha_{ins} \log \pi_\vartheta \left( \bm{\tilde{a}}_t^{ins} | \bm{h}_t^{ins} \right)  -  \nonumber \\
	& \alpha_{ins} \bar{\mathcal{H}}_{ins}  \bigg],
	\label{sac_s2p}
\end{align}

\noindent where $\alpha_{loc}$ and $\alpha_{ins}$ are temperature coefficients for entropy regularization of corresponding policies. $\bar{\mathcal{H}}_{loc}$ and $\bar{\mathcal{H}}_{ins}$ are entropy targets for different primitives. The critic loss of S2P-SAC can also be obtained via Eq.~\ref{new_drqv2_critic_loss}. Fig.~\ref{rr_sac} illustrates the training process of S2P-SAC on challenging tasks, where S2P is not able to enhance the exploration of SAC to find the optimal full insertion state with the force penalty. In addition, S2P-SAC performs slightly better than MAPLE on the X-shape task. These ablation results further show that S2P can enhance the performance of visual RL algorithms under severe constraints but cannot compensate for the deficiencies of visual RL algorithms on exploration.

\section{CONCLUSION}
This paper presents S2P based on visual reinforcement learning. S2P utilizes an ensemble of actors and critics to learn location and insertion actions, which are executed sequentially and can be generalized to one action response to the environment. S2P shows improved sample efficiency and success rates across different insertion tasks and can even explore the optimal state with force penalty. Physical experiments are performed to further validate the feasibility of S2P. The ablation study indicates that S2P fails to perform well on SAC compared to DrQ-v2. 

Future work will explore the greater potential of motion primitives to alleviate the dependence on underlying visual RL algorithms. Different primitives can be combined in more flexible and efficient ways by leveraging various modalities to explore better performance. The transfer performance of S2P can be further explored and enhanced in scenarios with uncertain orientation errors.

\appendix\label{appendix}
\section{Hyperparameters}\label{app:hyperparameters}

\begin{table}[h]
        \centering
        \caption{Hyperparameters used for training in simulation.}
        \renewcommand\tabcolsep{24.0pt}
        \begin{tabular}{@{}p{3cm}@{}p{4.5cm}@{}}
        \toprule[0.5mm]
        Hyperparameters    & Value     \\ 
        \hline
        Image size        & 96 $\times$ 96 $\times$ 3  \\
        Image augmentation & Random Shift (pad=4) \\
        Frame stack & 3   \\
        Discount factor $\gamma$               & 0.99            \\
        Replay Buffer size                      & 1e6            \\
        Batch size & 256 \\
        $N$-step return  &  3 \\
        Optimizer & Adam   \\
        Learning Rate & 1e-4 \\
        UTD ratio & 2 \\
        Max episode steps & 200 \\
        Linear decaying schedule &  linear $\left(1.0, 0.1, 30000 \right)$ \\
        Action repeat  & 1 (S2P) / 2 (DrQ-v2, MAPLE) \\
        CNN architecture & Conv (c=[32, 64, 128, 256], s=2, p=1) \\ 
        MLP architecture & Linear (c=[64, 1024, 1024], bias=True) \\
        $w_r, w_a$, $w_i$ & 1, 1, 70 (RJ45) / 40 (Others) \\
        $\lambda, \varepsilon _a, \delta _a, \varepsilon _f$ & 10, 5 mm, 0.05 rad, 1 mm \\
		Temperature coefficients & 0.1 (S2P-SAC and MAPLE) \\
		One-hot factor & 0.5 (MAPLE) \\
        \hline
        \end{tabular}
        \label{table: common_hyper}
\end{table}

\begin{table}[h]
    \centering	
    \caption{Specialized hyperparameters used for training tasks in Fig.~\ref{rr_force}}
    \label{table: special_hyper}
	\renewcommand\tabcolsep{24.0pt}
	\begin{tabular}{@{}p{3cm}@{}p{4.5cm}@{}}
	\toprule[0.5mm]
	Hyperparameters    & Value     \\ 
	\hline
	Force penalty weight   & $-0.8$ (X-shape) / $-0.5$ (V-shape)  \\
	Force penalty threshold & $15$ N \\
	Noise clip intervals & $\epsilon_{loc}, \epsilon_{ins}=0.3$ (S2P-DrQ-v2) / 0.3 (DrQ-v2) \\
	Action repeat & 1 \\
	\hline
	\end{tabular}
\end{table}

\bibliographystyle{IEEEtran}

\end{document}